\documentclass[10pt,twocolumn,letterpaper]{article}

\usepackage{iccv}
\usepackage{times}
\usepackage{epsfig}
\usepackage{graphicx}
\usepackage{amsmath}
\usepackage{amssymb}
\usepackage{xcolor}
\usepackage{eucal}

\usepackage[breaklinks=true,bookmarks=false]{hyperref}

\iccvfinalcopy 

\def\iccvPaperID{****} 

\ificcvfinal\fi

\begin{document}

\title{ScanERU: Interactive 3D Visual Grounding based on \\ Embodied Reference Understanding}

\author{Ziyang Lu
\and
Yunqiang Pei
\and 
Guoqing Wang*
\and
Yang Yang
\and
Zheng Wang
\and 
Heng Tao Shen
\\
School of Computer Science and Engineering, \\ University of Electronic Science and Technology of China\\
{\tt\small498358329@qq.com simon1059770342@foxmail.com gqwang0420@hotmail.com}
}

\maketitle
\ificcvfinal\thispagestyle{empty}\fi

\begin{abstract}
   {Aiming to link natural language descriptions to specific regions in a 3D scene represented as 3D point clouds, 3D visual grounding is a very fundamental task for human-robot interaction. 
   The recognition errors can significantly impact the overall accuracy and then degrade the operation of AI systems. 
   Despite their effectiveness, existing methods suffer from the difficulty of low recognition accuracy in cases of multiple adjacent objects with similar appearances.
   To address this issue, this work intuitively introduces the human-robot interaction as a cue to facilitate the development of 3D visual grounding. 
   Specifically, a new task termed Embodied Reference Understanding (ERU) is first designed for this concern. 
   Then a new dataset called ScanERU is constructed to evaluate the effectiveness of this idea. 
   Different from existing datasets, our ScanERU is the first to cover semi-synthetic scene integration with textual, real-world visual, and synthetic gestural information. 
   Additionally, this paper formulates a heuristic framework based on attention mechanisms and human body movements to enlighten the research of ERU. 
   Experimental results demonstrate the superiority of the proposed method, especially in the recognition of multiple identical objects. 
   Our codes and dataset \footnote[1]{Our project page: https://github.com/MrLearnedToad/ScanERU} are ready to be available publicly. }
   
\end{abstract}

\section{Introduction}

\begin{figure*}[tb]
 \centering
  \includegraphics[width=\textwidth]{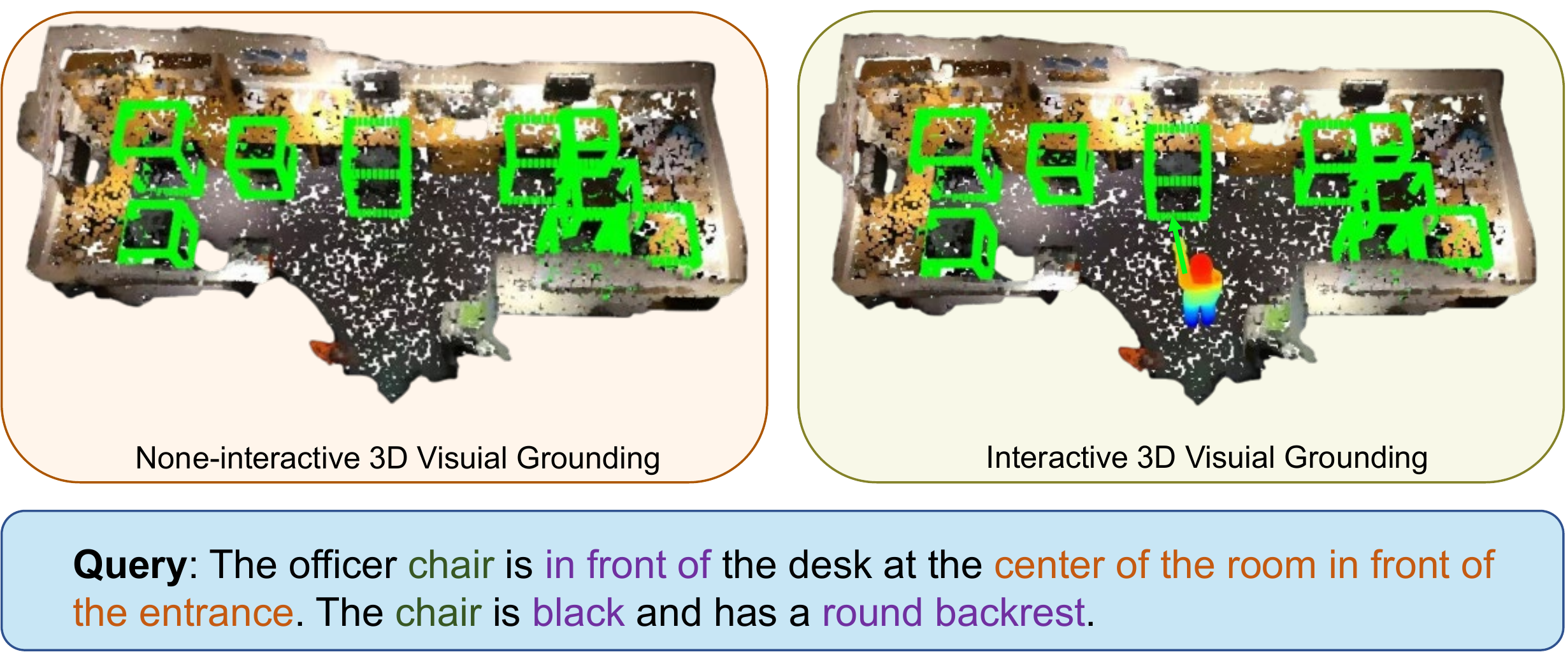}
 \caption{The comparisons between non-interactive and interactive visual grounding. The image displays a multitude of chairs in an office setting, presenting a challenge in identifying a unique referent among visually similar objects. Nevertheless, the inclusion of gestural information conveyed by a human agent can aid in the localization of the referred object by robots and AI systems.}
 \label{fig:first}
\end{figure*}

The ability to understand and localize objects from natural expression is critical for {the operation of AI systems}. 
{To this end, both 2D~\cite{deng2018visual,deng2021transvg, yang2022improving, qiao2020referring, wang2019neighbourhood} and 3D~\cite{huang2022multi, chen2020scanrefer, yang2021sat, huang2021text, yuan2021instancerefer, LichenZhao20213DVGTransformerRM} visual groundings refer} to understanding how words and language can be linked to visual information in images and videos, and how this information can be used to recognize and understand objects, scenes, and actions in the environment. 
{The former takes 2D images or video frames as input and suffers from the limitation of fully localize objects and impractical problems in the real applications of robots and virtual/augmented reality~\cite{chen2020scanrefer}, which is caused by the restricted nature of 2D images.}
{Therefore, the research of 3D visual grounding based on 3D point could data has emerged, which aims to locate a specific object or region in a 3D scene referred by a natural language description.} 
For example, given a point cloud of a living room and a query such as ``the blue sofa near the window", the goal is to identify and highlight the corresponding sofa in the scene. 
{
Due to the compatibility with 3D data used in Simultaneous Localization and Mapping (SLAM)~\cite{thrun2005multi, taketomi2017visual} technology, 3D visual grounding can provide a more accurate and precise mapping of the environment and localization of the robot. 
Additionally, because of 3D point cloud data, it can describe the objects in the environment in more detail, such as their shape and size. 
These properties guarantee its wide range of applications, including robotics, augmented reality, virtual reality, and human-robot interaction \cite{yang2021sat, huang2022multi}, where natural language can be used as an intuitive and flexible way to interact with 3D environments.
}
{Several works have studied this problem with various methods on different datasets. 
For example, Chen \etal \cite{chen2020scanrefer} presented the ScanRefer dataset and a comprehensive end-to-end framework based on 3D point clouds. 
Achlioptas \etal \cite{PanosAchlioptas2020ReferIt3DNL} introduced another dataset of ReferIt3D that focuses on identifying objects among instances of the same fine-grained category. 
Wang \etal \cite{liu2021refer} proposed Refer-it-in-RGBD, a bottom-up approach for 3D visual grounding in RGBD images that does not require scene reconstruction.}
{
Despite the success of existing work, identifying the referred object among multiple adjacent objects with similar appearances is still a great challenge in this task. 
The incorrect identification or low accuracy of recognition will seriously affect the application and deployment of the visual grounding.  
}
To address the challenge, existing methods have proposed various techniques such as attention mechanisms, transformers \cite{LichenZhao20213DVGTransformerRM, roh2022languagerefer}, semantics-assisted training \cite{yang2021sat}, multi-view fusion \cite{huang2022multi}, etc. 

{However, the explorations on this concern are clearly insufficient. 
Differently, this work solves this issue from a new perspective of human-robot interaction, which can act as rich cues to promote the recognition of visually similar objects in 3D space. 
A similar idea that language and gestural coordination is important in 2D visual grounding was confirmed by chen \etal~\cite{YixinChen2021YouRefItER}. 
Specifically, they constructed the YouRefIt dataset, mainly consisting of videos, where humans jointly leverage language and gestures to refer to objects, and thus significantly improved the accuracy of 2D visual grounding by the incorporation of gesture information. 
Inspired by their success, this work aims to further improve the accuracy of 3D visual grounding on multiple adjacent objects with similar appearances by incorporating human gestures to disambiguate referring expressions and accurately identify the referred object. 
Specifically,  we are the first to design a new task for 3D visual grounding termed Embodied Reference Understanding (ERU), which is built upon the embodied perspective of the agent. 
To better evaluate such a task, a new dataset called ScanERU is constructed based on existing datasets by incorporating textual, real-world visual, and synthetic gestural information into semi-synthetic scenes. 
Finally, to validate the effectiveness, we formulate a heuristic framework based on attention mechanisms and human body movements. 
Different from prior work, this work incorporates considerations for interactions with other intelligent agents, which thus provides a more natural way of human-robot interaction and a more human-like understanding of the 3D world. 
}

In conclusion, our contributions are as follows:

\begin{enumerate}
    \item \textbf{A novel task} called Embodied Reference Understanding (ERU) for 3D visual grounding is designed, which first jointly leverages language and gestures to refer to objects in 3D point clouds.
    \item \textbf{A new dataset} called ScanERU is constructed, which covers diverse and challenging semi-synthetic scenearios with synthetic gestural information. 
    \item \textbf{A heuristic framework} based on attention mechanisms and human body movements is proposed to evaluate our effectiveness on the recognition of multiple identical objects or complex spatial relations. 
\end{enumerate}

\section{Related Work}
\subsection{Non-Interactive Visual Grounding}
2D visual grounding is the process of determining the most relevant object or region in an image through a natural language query. There are several datasets available for this task, encompassing both real world \cite{kazemzadeh2014referitgame,yu2016modeling,mao2016generation,plummer2015flickr30k,HarmdeVries2016GuessWhatVO} and synthetic world \cite{RuntaoLiu2019CLEVRRefDV} scenarios. The conventional methods encompass two-stage \cite{mao2016generation,yu2016modeling} and one-stage \cite{ZhengyuanYang2019AFA,ZhengyuanYang2020ImprovingOV} approaches. This field has undergone extensive study and various methods \cite{yu2018mattnet,liu2019learning} and have been proposed to improve its accuracy, including the implementation of transformer-based frameworks \cite{yu2018mattnet} that construct text-conditioned discriminative features. Visual features play a crucial role in visual grounding as they assist in identifying objects and regions in an image. However, 2D images only provide a single perspective, while a 3D representation can be viewed from multiple angles, thereby providing a more comprehensive understanding of the objects and their relationships. Additionally, 2D images can frequently be ambiguous, leading to multiple possible interpretations or outcomes. In contrast to 2D visual grounding, 3D visual grounding involves establishing a connection between language and 3D objects in a point cloud environment, allowing the model to capture spatial features. The objective is to align natural language descriptions with the corresponding 3D objects and their attributes in a 3D setting. Chen \etal \cite{chen2020scanrefer} presented the ScanRefer Dataset and a comprehensive end-to-end framework for the visual grounding task. The ReferIt3D \cite{PanosAchlioptas2020ReferIt3DNL} introduced two datasets, similar to the ScanRefer dataset, but with a focus on identifying the referred object among instances of the same fine-grained category. Most approaches\cite{chen2020scanrefer,LichenZhao20213DVGTransformerRM,JiamingChen2022HAMHA,yuan2021instancerefer,bakr2022look,huang2022multi} adopt the two-stage framework established by ScanRefer, while the 3D-SPS method \cite{JunyuLuo20223DSPSS3} devises a single-stage solution to the task. Cai \etal \cite{cai20223djcg} developed a unified joint framework that accommodates both the grounding task and captioning task. Inspired by the transformer, recent works such as 3DVG-Transformer \cite{LichenZhao20213DVGTransformerRM} and SAT \cite{yang2021sat} have integrated attention mechanisms into the framework. The most recent work, HAM \cite{JiamingChen2022HAMHA}, leverages the spatially-global and spatially-local attention to locate referred objects, achieving the best result of Acc@0.5 on the ScanRefer Challenge. However, the sparse, noisy, and limited semantic information of point clouds compared to 2D images make it difficult to accurately locate a referred object \cite{yang2021sat}. Additionally, the proximity of the referent to adjacent objects in the scene can lead to localization errors \cite{bakr2022look, PanosAchlioptas2020ReferIt3DNL, chen2020scanrefer, LichenZhao20213DVGTransformerRM}, and view-dependent descriptions can result in poor localization performance for referent localization based on spatial terms \cite{huang2022multi, yang2021sat, LichenZhao20213DVGTransformerRM, huang2021text, he2021transrefer3d, yuan2021instancerefer, feng2021free}. There are also localization errors when locating a unique referent among multiple visually similar objects \cite{bakr2022look, JunyuLuo20223DSPSS3, huang2022multi, yang2021sat, LichenZhao20213DVGTransformerRM, huang2021text, he2021transrefer3d, yuan2021instancerefer, feng2021free, liu2021refer, PanosAchlioptas2020ReferIt3DNL, chen2020scanrefer}. Our approach introduces a new task of 3D visual grounding in a human-in-the-loop-based scenario, where body gestures are integrated into the scene to mitigate localization errors resulting from sparse, noisy, and semantically limited point clouds, object proximity, difficulty in distinguishing a unique referent among visually similar objects, and view-dependent descriptions.

\begin{figure}[tb]
 \centering
  \includegraphics[width=\columnwidth]{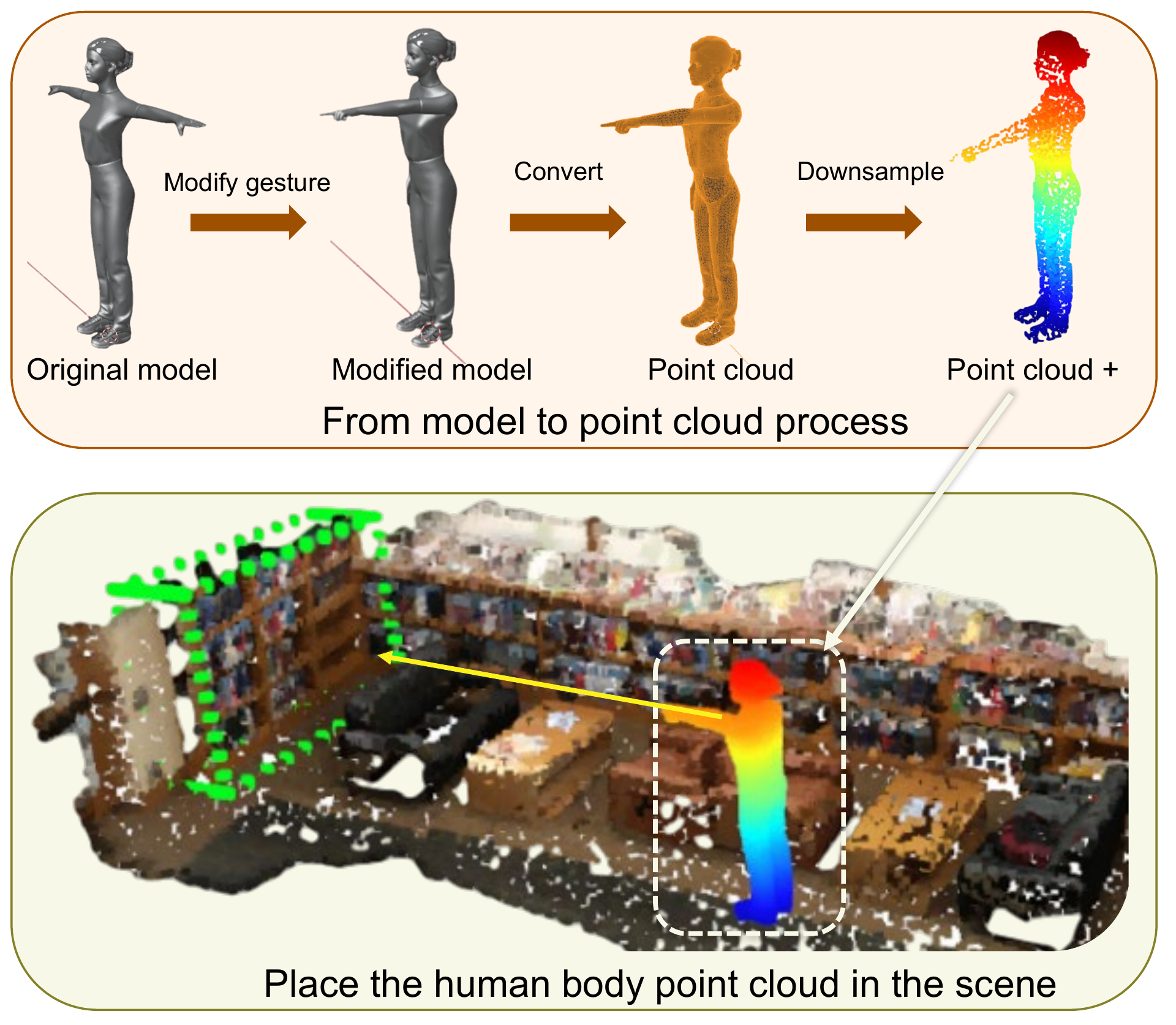}
 \caption{Generation procedure of human point cloud.}
 \label{fig:humanpoint}
\end{figure}

\subsection{Interactive Visual Grounding}

In order to improve the accuracy and facilitate a more natural method of communication between humans and agents, Chen \etal \cite{YixinChen2021YouRefItER} introduced the ERU (Embodied Reference Understanding) task, in which an agent uses both language and gestures to refer to an object to another agent in a shared physical environment. This is accompanied by the introduction of the YouReferIt dataset, which is a 2D multi-modal dataset encompassing textual, visual, and gestural information. Building on psychological studies of human pointing gestures, Li \etal \cite{YangLi2022UnderstandingER} proposed a new architecture using the virtual touch line (a line connecting the eye and the fingertip) and a transformer, leading to a significant improvement in the ERU task. Our work expands the ERU task to point cloud environment and studies the disambiguation effect of human gesture. \autoref{fig:first} shows the comparisons between non-interactive and interactive visual grounding. 

Interactive visual grounding is a task that involves using natural language and gestures to refer to objects or regions in an image or a 3D scene. For example, a human may point to a shelf and say ``the blue bottle next to the green one” to indicate an object to a robot. It aims to improve the accuracy and robustness of visual grounding by incorporating human-robot interaction as a cue, such as asking questions or requesting feedback from the human. For example, a robot may ask “do you mean this one?” and show an object to the human for confirmation \cite{ZhengyuanYang2020ImprovingOV}. Interactive visual grounding can enhance human-robot communication and collaboration by allowing robots to understand human references more accurately and efficiently, especially in cluttered or ambiguous environments where multiple objects may look similar or occlude each other.

\section{Dataset}
To study ERU task in 3D environment, we propose the ScanERU dataset, a semi-synthetic dataset for ERU task. Our dataset is based on the ScanRefer \cite{chen2020scanrefer} and ScanNet \cite{AngelaDai2017ScanNetR3} datasets and includes 706 unique indoor scenes, 9,929 referred objects, and 46,173 descriptions. We synthesize human point cloud data and pointing gestures for each referred object. For the validity test of ScanERU, please refer to the supplement.

\subsection{Data Annotation}
The procedure of annotating the semi-synthetic dataset is conducted through a visualization UI and LabelImg, which presents the workers with both the point cloud and a top view of the scene with the non-referred objects (e.g., ceiling) faded out. To ensure the quality and accuracy of our annotations, our workers are instructed to annotate five possible positions of the synthetic agent on the top view. The annotations are then subjected to automatic checks to verify that there is no object obstructing the line of sight between the synthetic agent and the referred object. Upon successful completion of the checks, the verified annotations are assigned to each referred object in the dataset. To enhance the generalization of the modality information of ``finger pointing to the referred object", we modified Li's ``Virtual Touch Line" \cite{YangLi2022UnderstandingER} approach in the ERU2D domain, where the line from the eye to the fingertip points directly to the center of the referred object without any deviation. In our dataset generation method, we adjusted the angle fluctuation range of the arm movement by ±5° to ensure that the ray emitted by the gesture can pass through the referred object, rather than pointing directly at the object's center. The output of the annotation process are the index between point cloud of synthetic human agents and referred objects and the coordinate of the synthetic agent in the scene point cloud.

\subsection{Generation Procedure}

To ensure the diversity and variability of the synthetic gestural information, we utilize 10 different human models both male and female. The meshes are batch-processed using Blender. The detailed process is shown as \autoref{fig:humanpoint}:

\begin{itemize}
    \item \textbf{Rotate the skeleton to generate a character pool.} To generate a character pool, we apply a rotation transformation to the skeleton of each character. Specifically, we create a `pointing’ gesture for each character by rotating their hand and arm towards the referred object. The `pointing` gesture consists of two parts: the choice of left or right hand, and the rotation angle of the arm. We vary the rotation angle from -90 to +90 degrees with an interval of 0.5°. Moreover, we introduce random angle perturbation to ensure the diversity and realism of each synthetic agent. The randomly rotated skeleton includes the left or right arm (both upper and lower parts), the left or right hand, and the head. The perturbation angle is drawn from a Gaussian distribution with a range of -3° to +3°. 

    \item \textbf{Convert the mesh files of human models into point clouds.} We convert the mesh files of human models into point clouds and perform voxel down-sampling to align them with the ScanNet dataset. The voxel size is set to 0.25cm. Then, we use random down-sampling to reduce the number of points to 3000. If the number is below 3000, we pad zeros to fill up the remaining points.
    
    \item \textbf{Load the matching point cloud from the character pool into the scene.} Due to the large size of the point cloud file, we do not load it into the scene point cloud until training or evaluation time. The dataset is loaded along with label, vertex, and normal information based on the annotated position and the index between point cloud of synthetic human agents and referred objects. And each referred object is associated with 3-5 different synthetic agents pointing at it from different positions.
 \end{itemize}

\subsection{Dataset Statistics}

\begin{figure}[tb]
 \centering
  \includegraphics[width=\columnwidth]{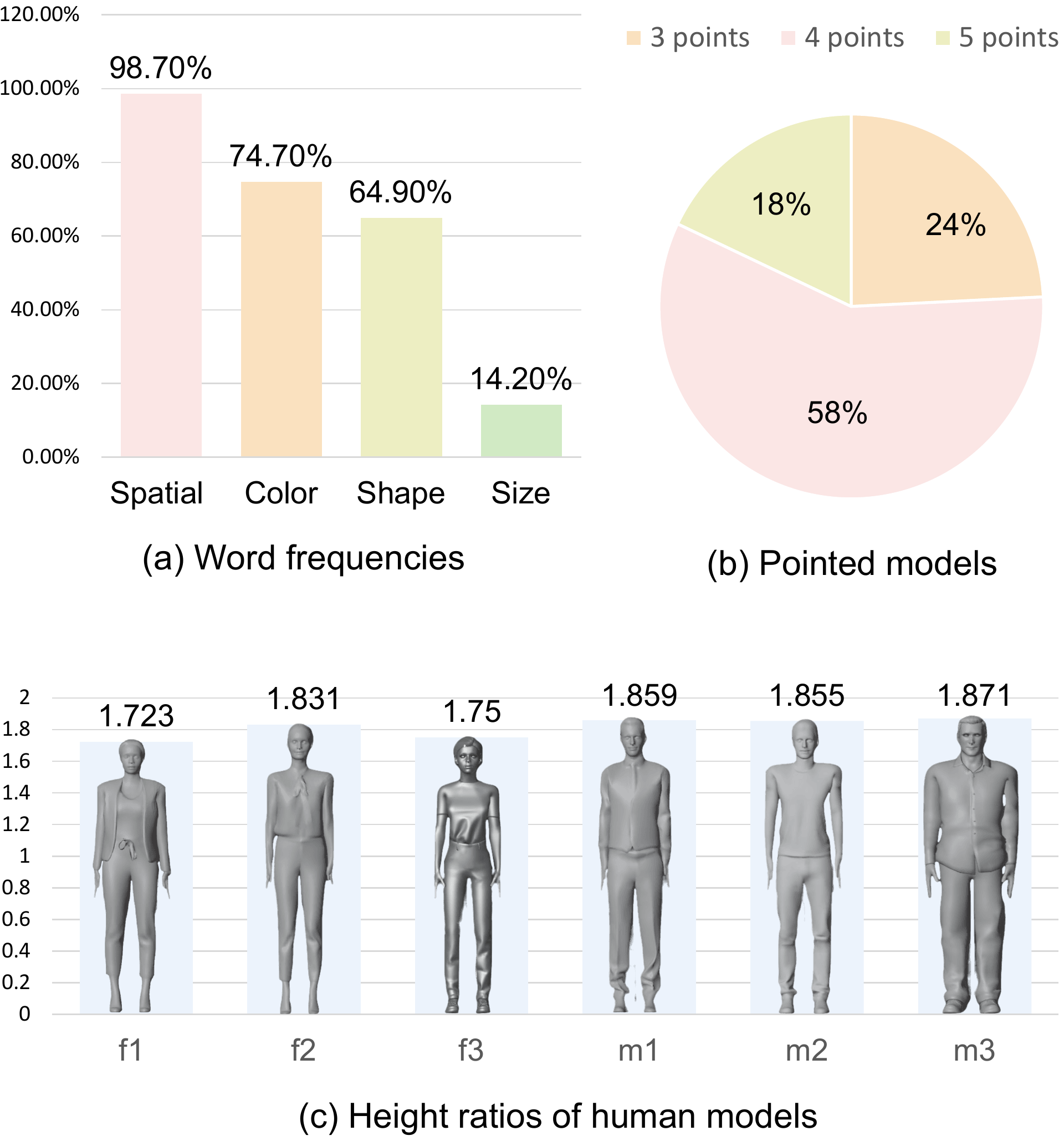}
 \caption{Dataset Statistics of ScanERU.}
 \label{fig:statistics}
\end{figure}

After filtering 46,173 descriptions in ScanRefer \cite{chen2020scanrefer} for 706 ScanNet \cite{AngelaDai2017ScanNetR3} scenes, we find that 41,034 mentioned object attributes, such as color, shape, and size. These attributes are used to describe over 250 types of common indoor objects, resulting in complex and diverse descriptions. Spatial language (98.7$\%$), color (74.7$\%$), and shape terms (64.9$\%$) are frequently used, while size information is only conveyed in 14.2$\%$ of the descriptions (see \autoref{fig:statistics}(a)). The complexity of the descriptions poses a challenge in distinguishing between referred objects and neighboring ones.

\begin{figure*}[tb]
 \centering
  \includegraphics[width=\textwidth]{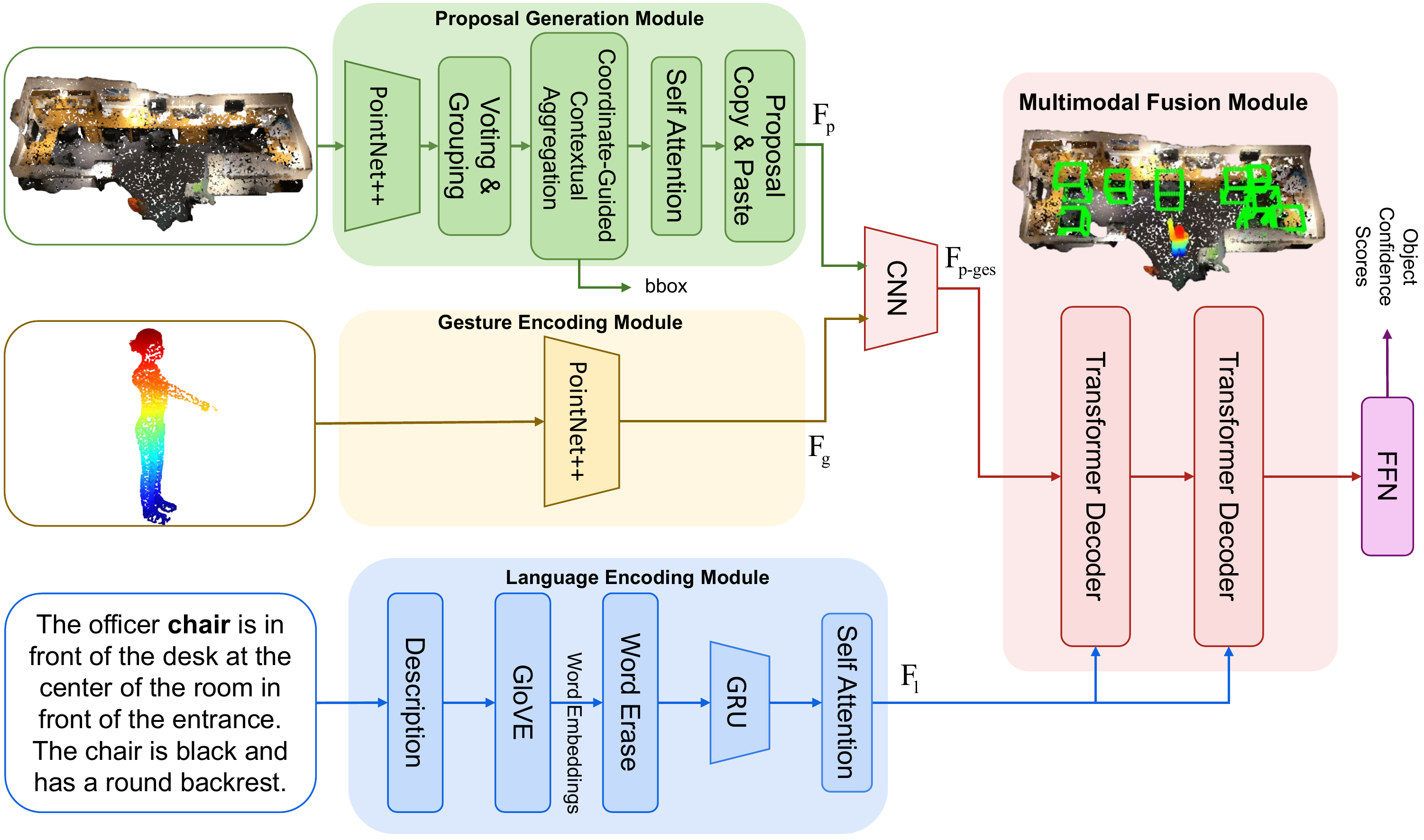}
 \caption{The architecture of the proposed ScanERU. 
 It takes a point cloud of the scene, a point cloud of a human agent, and a description of the referred object as input, through the processing of the proposal generation module, the gesture extraction module, and the language module to extract features of different modalities. 
 Based on a multi-modal fusion module, it outputs the confidence scores of the bounding boxes. 
 The highest confidence score is the final prediction. Best viewed in color.}
 \label{fig:network}
\end{figure*}

Our dataset encompasses ten distinct original human models, comprising of three adult males, three adult females, one male child, one female child, one elderly male, and one elderly female. We generated 7,200 different point clouds of the human agent from these models. Half of the dataset involves left-hand pointing gestures, while the remaining half utilizes right-hand pointing gestures. On average, the height of the characters is 175.7 cm. Each referred object is pointed at from three to five positions, with an average of 3.93 positions.
\autoref{fig:statistics}(b) counts the number of models in the scene that are pointed by different numbers of human models.
Our dataset reveals intriguing human-in-the-loop phenomena, including gesture-assisted reference and human-agent interactions.
\autoref{fig:statistics}(c) shows six human body models (three males and three females, used to differentiate the height of male and female models, but since the heights of boy and girl models are similar, they are not shown here), and lines them up according to their respective height ratios.
For more detailed statistics, please refer to the supplement.

\section{Methodology}
{This section describes our work in detail. Sec 4.1 gives an overview of the ScanERU framework. Sec 4.2 explains how we generate proposals, encode gestures, and encode language. Sec 4.3 presents how we fuse multi-modal features. Sec 4.4 defines the loss function.}

\subsection{Overview}

Shown as \autoref{fig:first}, the ScanERU comprises three inputs: the point cloud of the entire 3D scene, a description of the referred object, and the point cloud of the human agent. The scene point cloud $P_p \in \mathbb{R}^{N\times (3+K)}$ contains $N$ points' coordinate and $K$-dimensional features such as RGB and normal vectors. The description is tokenized and transformed into word embeddings using the GloVE \cite{pennington2014glove} model. The human agent point cloud is similar to the scene point cloud, except that its features only include normal vectors. The objective of the task is to locate the referred object and output its bounding-box in world coordinates.

The ScanERU framework consists of four modules: proposal generation, gestural encoding, language encoding, and multi-modal fusion. To better leverage the features among language, gesture, and the scene point cloud, an attention mechanism \cite{AshishVaswani2017AttentionIA} is employed in our work. The proposal generation module is the same as that used in 3DVG-Transformer \cite{LichenZhao20213DVGTransformerRM} and it generates a bounding-box from object proposal while extracting context-aware features. The proposal features are represented by $F_p\in \mathbb{R}^{M\times H}$, for $M$ proposals with $H$-dimensional features. The gestural encoding module uses a PointNet++ \cite{CharlesRQi2017PointNetDH} to extract the features $F_g \in \mathbb{R}^{M\times H}$ of the human agent's point cloud. Similar to ScanRefer \cite{chen2020scanrefer} and 3DVG-Transformer \cite{LichenZhao20213DVGTransformerRM}, the language encoding module aggregates the word embeddings into the language features $F_l\in \mathbb{R}^{L\times H}$ and global language features using a GRU \cite{JunyoungChung2014EmpiricalEO} cell and a self-attention module. The multi-modal fusion module leverages attention mechanism \cite{AshishVaswani2017AttentionIA} to fuse proposal features $F_p$, gestural features $F_g$, and word features $F_l$, thereby generating the confidence score of each bounding-box. Specifically, this study centers on the combination of gestural information with the proposal and word information, aiming to disambiguate referring expressions and accurately identify the referred object.

\subsection{Feature Encoding Modules}
\textbf{Proposal generation module.} Our proposed method utilizes a PointNet++ \cite{CharlesRQi2017PointNetDH} backbone and a voting and grouping module \cite{qi2019deep}, similar to ScanRefer \cite{chen2020scanrefer} and 3DVG-Transformer \cite{LichenZhao20213DVGTransformerRM}, to process the point cloud of the scene and group them into individual clusters. Subsequently, we employ the coordinate-guided contextual aggregation (CCA) module, as utilized in 3DVG-Transformer \cite{LichenZhao20213DVGTransformerRM}, to generate refined proposal features as $F_{p0}$ and bounding-boxes. To further refine the proposal features, a self-attention module is applied, which takes the refined proposal features $F_{p0}$ as input and studies the contextual relationships within the refined proposal features $F_{p0}$. In addition, we employ a copy\&paste module, akin to the method used in 3DVG-Transformer \cite{LichenZhao20213DVGTransformerRM}, to leverage the over-fitting issue, producing the output as $F_{p}$.

\textbf{Gesture encoding module.} The point cloud of the human agent is also processed with a PointNet++ \cite{CharlesRQi2017PointNetDH} backbone, similar to the point cloud of the scene. We use the PointNet++ \cite{CharlesRQi2017PointNetDH} to obtain the gestural features of the $F_g$. 

\textbf{Language encoding module.} The textual description input is encoded using the GloVe \cite{pennington2014glove} and GRU \cite{JunyoungChung2014EmpiricalEO} module, which is the same module used in the ScanRefer \cite{chen2020scanrefer} framework. Moreover, our training process is augmented with the word erase training strategy, which has been shown to be beneficial in 3DVG-Transformer \cite{LichenZhao20213DVGTransformerRM}. Furthermore, to refine the language features, we employ a self-attention module to generate $F_l$ from the GRU output.

\subsection{Multi-Modal Fusion Module}

\begin{figure}[tb]
 \centering
  \includegraphics[width=\columnwidth]{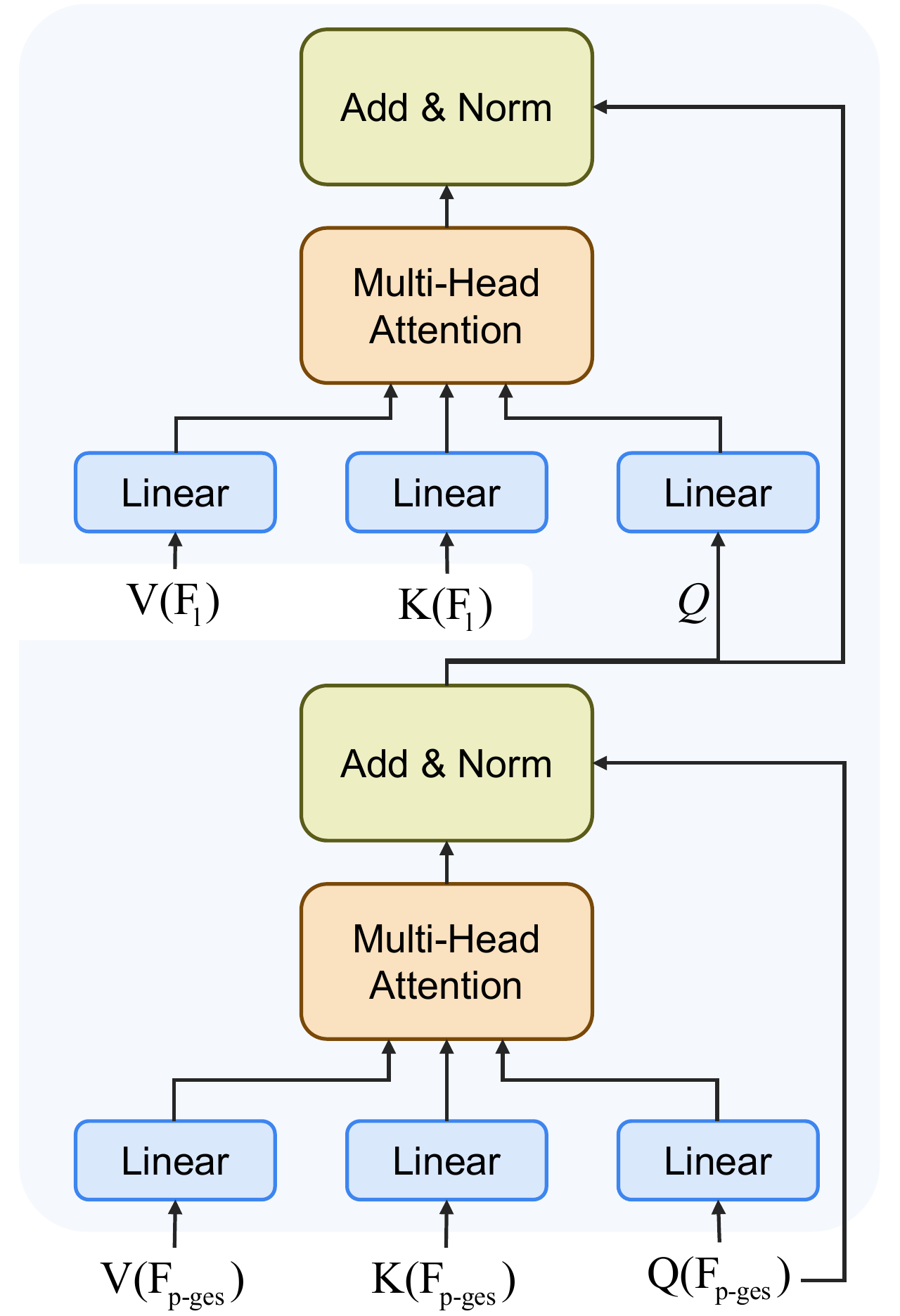}
 \caption{The network structure of the transformer decoder, which is utilized in our approach involves the vanilla version described in \cite{AshishVaswani2017AttentionIA}. This design allows us to effectively explore the correlations between gesture features, proposal features, and language features.}
 \label{fig:decoder}
\end{figure}

As illustrated in \autoref{fig:network}, the proposal features $F_p$ and gestural features $F_g$ are concatenated and fused using a convolution block, with the resulting features being denoted as $F_{p-ges}$. Subsequently, we use a 2-layer stacked transformer decoder to exploit the relationship of proposal-gestural features $F_{p-ges}$ and language features $F_l$, where the proposal-gestural features $F_{p-ges}$ serves as query while the language features $F_l$ serves as key and value. The detailed architecture of the transformer decoder is shown in \autoref{fig:decoder}. Finally, the output of the stacked transformer decoder is fed into a feed-forward network (FFN) layer and a softmax activation layer to generate the confidence score of each bounding-box.

\begin{figure*}[tb]
 \centering
  \includegraphics[width=\textwidth]{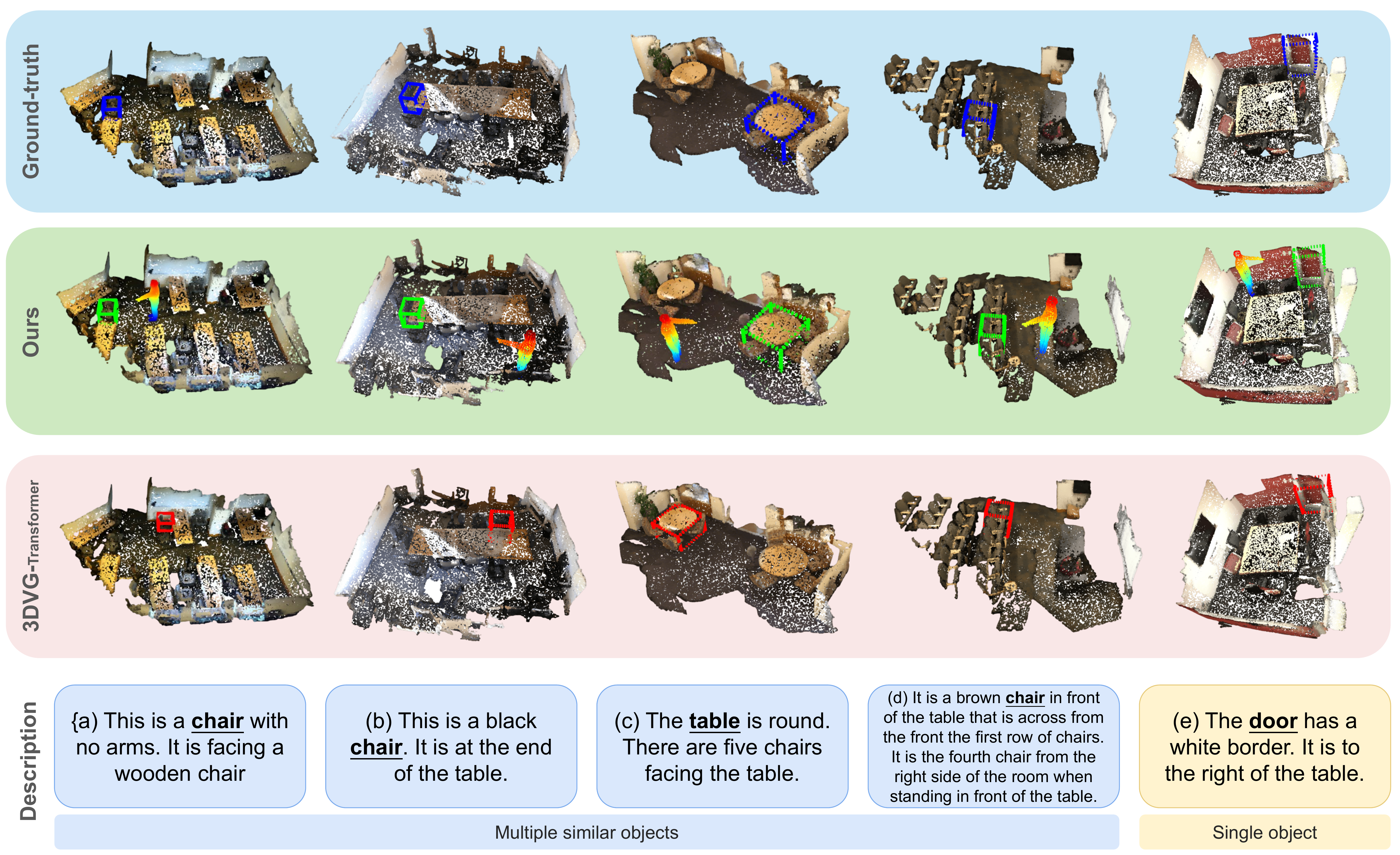}
 \caption{Qualitative results from 3DVG-Transformer\cite{LichenZhao20213DVGTransformerRM} and our ScanERU. The GT boxes are marked in blue. If one predicted box has an IoU score higher than 0.5, this box is marked in green, otherwise it is marked in red.}
 \label{fig:qualitative}
\end{figure*}

\begin{table*}[tb]
  \caption{Comparasion of visual grounding performances on ScanRefer and ScanERU dataset.}
  \label{tab:trust}
	\centering%
\resizebox{\linewidth}{!}{ 
\begin{tabular}{ccccccccc}
\hline
Methods              & Venue    & \multicolumn{1}{c|}{Modality} & \multicolumn{2}{c}{Unique}            & \multicolumn{2}{c}{Multiple}          & \multicolumn{2}{c}{Overall}           \\
                     &          & \multicolumn{1}{c|}{}         & Acc@0.25 & Acc@0.5 & Acc@0.25 & Acc@0.5 & Acc@0.25 & Acc@0.5 \\ \hline
\multicolumn{9}{c}{Results on the ScanRefer validation set}                                                                    \\ \hline
ScanRefer \cite{chen2020scanrefer}            & ECCV2020 & \multicolumn{1}{c|}{2D+3D}    & 76.33    & 53.51   & 32.73    & 21.11   & 41.19    & 27.40    \\
TGNN \cite{huang2021text}                 & ICCV2021 & \multicolumn{1}{c|}{2D+3D}    & 68.61    & 56.80    & 29.84    & 23.18   & 37.37    & 29.70    \\
InstanceRefer \cite{yuan2021instancerefer}        & ICCV2022 & \multicolumn{1}{c|}{3D}       & 77.45    & 66.83   & 31.27    & 24.77   & 40.23    & 32.93   \\
3DVG-Transformer \cite{LichenZhao20213DVGTransformerRM}     & ICCV2021 & \multicolumn{1}{c|}{2D + 3D}  & 81.93    & 60.64   & 39.30     & 28.42   & 47.57    & 34.67   \\
3DJCG \cite{cai20223djcg}                & CVPR2022 & \multicolumn{1}{c|}{2D + 3D}  & 83.47    & 64.34   & 41.39    & 30.82   & 49.56    & 37.33   \\
3D-SPS \cite{JunyuLuo20223DSPSS3}               & CVPR2022 & \multicolumn{1}{c|}{2D + 3D}  & \textbf{84.12}    & 66.72   & 40.32    & 29.82   & 48.82    & 36.98   \\ 
HAM \cite{JiamingChen2022HAMHA}               &  & \multicolumn{1}{c|}{3D}  & 79.24    & \textbf{67.86}   & 41.46    & 34.03   & 48.79    & 40.06   \\ \hline
\multicolumn{9}{c}{Results on the ScanERU validation set}                                                                      \\ \hline
ScanERU              &          & \multicolumn{1}{c|}{3D}       & 81.26         & 61.91        & \textbf{47.99}         & \textbf{35.89}        & \textbf{54.45}         & \textbf{40.94}       \\
\hline
\end{tabular}
}
\end{table*}

\subsection{Loss Function}
In our approach, we employ a loss function similar to that used in 3DVG-Transformer \cite{LichenZhao20213DVGTransformerRM}, which is represented as $L=0.3L_{loc}+10L_{det}+0.1L_{cls}$. Here, $L_{loc}$  denotes the localization loss, $L_{det}$  represents the object detection loss, and $L_{cls}$  indicates the language-to-object classification loss. Furthermore, we can decompose $L_{det}$ as
$
L_{det} = L_{vote-reg} + 0.1L_{objn-cls} + 0.1L_{sem-cls} + L_{box}
$
where $L_{vote-reg}$ is the vote regression loss, $L_{objn-cls}$ and $L_{sem-cls}$ are the objectness and semantic classification losses, respectively, and $L_{box}$  denotes the bounding-box loss. The bounding-box loss can be further decomposed as
$
  L_{box} = L_{center-reg} + 0.1L_{size-cls} +L_{size-reg}
$
where $L_{center-reg}$ and $L_{size-reg}$ are the center and size regression losses, respectively, and $L_{size-cls}$ denotes the size classification loss.

\section{Experiment}
\textbf{Dataset Split.} In our experimental evaluation, we conduct tests on both the ScanRefer \cite{chen2020scanrefer} dataset and the ScanERU dataset. Following the same protocol as the ScanRefer dataset \cite{chen2020scanrefer}, we split it into train, validation, and test sets with 36,665, 9,508, and 5,410 samples, respectively. Similarly, we split the ScanERU dataset into train and validation sets using the same ratio as that of the ScanRefer dataset \cite{chen2020scanrefer}, where the number of samples in the training and validation sets is 36,665 and 9,508, respectively.

\textbf{Baseline.} We devise the baselines by comparing our method with state-of-the-art methods on 3D visual grounding task.

\textbf{Metric.} Following the standard evaluation metric for 3D visual grounding tasks, we employ two commonly used metrics, namely Acc@0.25IoU and Acc@0.5IoU, to measure the performance of our method. Additionally, we also report the ``unique," ``multiple," and ``overall" scores, as defined in the ScanRefer dataset \cite{chen2020scanrefer}. The ``unique" score measures the performance when there is only a single object of its class in the scene, whereas the ``multiple" score measures the performance when there are more than one similar object of its class in the scene. The ``overall" score is the weighted average of the ``unique" and ``multiple" scores.

\subsection{Quantitative Study}

In \autoref{tab:trust}, We compare the performance of our ScanERU method with several existing 3D visual grounding methods. Additionally, in the ScanRefer dataset \cite{chen2020scanrefer}, the modality ``3D" indicates that the input includes only coordinates, RGB, and normal vectors, whereas ``2D+3D" indicates that an additional multiview 2D image is included as input. In the ScanERU dataset, there is only 3D modality. Since our ScanERU entries and ScanRef entries are the same in terms of 3D scenes and text descriptions, the results obtained by other methods through ScanRef training or ScanERU training are consistent. At the same time, we classified the methods using ScanRef and ScanERU in \autoref{tab:trust}.

In the ``multiple" subset, our proposed method exhibits superior performance compared to the state-of-the-art (SOTA) method, with an improvement of 6.5$\%$ for Acc@0.25 and 1.8$\%$ for Acc@0.5, when the SOTA is trained and evaluated on the ScanRefer dataset. Moreover, our method's overall accuracy surpasses the SOTA method due to its enhanced disambiguation ability. Remarkably, our proposed method is the sole approach that attains an accuracy surpassing 50$\%$ at 0.25 IoU, particularly without relying on the use of auxiliary 2D images, which underscores the effectiveness of incorporating human gestures in localizing and distinguishing multiple similar objects. These results validate our proposed method's efficacy in improving localization performance in complex indoor environments.

\subsection{Qualitive Study }

\autoref{fig:qualitative} depicts a visualization of the performance of our proposed method and the baseline 3DVG-Transformer \cite{LichenZhao20213DVGTransformerRM}. The ground-truth bounding boxes are denoted in blue, whereas the predicted boxes are highlighted in green if their IoU score with the ground truth is above 50$\%$, and in red otherwise. The results demonstrate that our method is capable of successfully localizing the referred object in complex environments with multiple similar objects, while the baseline method exhibits failure cases. We identify two main causes of failure for 3DVG-Transformer \cite{LichenZhao20213DVGTransformerRM}. The first cause is the difficulty of distinguishing fine-grained features from point cloud data. As illustrated in \autoref{fig:qualitative} (a), the terms ``no arms" and ``wooden chair" are not sufficiently descriptive to differentiate the object from others, even for human observers. The second cause is the ambiguity or complexity of the description. In \autoref{fig:qualitative} (b), the phrase ``at the end of the table" is unclear, meaning that there are multiple possible objects being referred. These results indicate that the language-only modality has limitations in disambiguating the correct object, particularly in challenging environments.

\begin{table}[tb]
\caption{Ablation study on the ScanERU validation set. We only report the overall score under ``Acc@0.25" and ``Acc@0.5" subset. }
  \label{tab:ablation}
	\centering%
\begin{tabular}{c|ccc}
\hline
Methods        & Acc@0.25 & Acc@0.5  \\ \hline
Ours$_{ges-only}$  & 17.46       & 13.44          \\
Ours$_{lang-only}$ & 49.18       & 35.62          \\
Ours$_{full}$ & 54.45       & 40.94          \\ \hline
\end{tabular}
\end{table}

\subsection{Ablation Study}
In this subsection, we aim to conduct a detailed analysis of the contributions of textual and gestural modalities in our proposed approach.

\textbf{Contribution of textual information.} To investigate the impact of textual information, we conduct an experiment in which only the scene point cloud and human agent point cloud are used as input. The results in \autoref{tab:trust} demonstrate that the performance of Ours$_{ges-only}$ is significantly inferior to that of Ours$_{full}$, which suggests that without textual information, the localization of the referred object can be ambiguous. This is because there could be multiple objects pointed at by the human agent, and only textual information can disambiguate this ambiguity.

\textbf{Contribution of gestural information.} Ours$_{lang-only}$ is trained on the original ScanRefer dataset \cite{chen2020scanrefer}, with the input of scene point cloud and textual description. The results in \autoref{tab:ablation} show that Ours$_{full}$ achieves a significant improvement over Ours$_{lang-only}$, which highlights the crucial role of gestural cues in the ERU task. Therefore, incorporating gestural information can significantly help distinguish multiple similar objects and improve the localization performance of the model.

\section{Conclusion}
This paper introduces a new task called embodied reference understanding (ERU) in 3D point cloud environments. In ERU, an agent employs both language and gestures to refer to an object in a shared physical environment. To facilitate research in this area, we propose the ScanERU, a semi-synthetic  dataset. Furthermore, we provide a new framework for the ERU task in 3D environments that leverages multi-modal features and attention mechanisms. Our work outperforms the best 3D visual grounding methods, particularly in recognizing multiple identical objects. This research contributes significantly to the field of 3D visual grounding by introducing human gestures as an additional modality that can help disambiguate referring expressions and accurately identify referred objects. Additionally, our work highlights the importance of embodied perspective and human-agent interaction for achieving a more natural and human-like understanding of the 3D world. For future work, we plan to expand our dataset to include more diverse scenes and gestures and explore other modalities such as voice or eye gaze to further enhance ERU performance.

\newpage
{\small
\bibliographystyle{ieee_fullname}
\bibliography{ours}
}

\end{document}



\title{ScanERU: Interactive 3D Visual Grounding based on \\ Embodied Reference Understanding \\ \large{\textmd{Supplementary Material}}}
\

\author{Ziyang Lu
\and
Yunqiang Pei
\and 
Guoqing Wang*
\and
Yang Yang
\and
Zheng Wang
\and 
Heng Tao Shen
\\
School of Computer Science and Engineering, \\ University of Electronic Science and Technology of China\\
{\tt\small498358329@qq.com simon1059770342@foxmail.com gqwang0420@hotmail.com}
}


\maketitle
\ificcvfinal\thispagestyle{empty}\fi

\renewcommand{\thesection}{\Alph{section}}
\section{Validity Test of ScanERU Dataset}
To evaluate the effectiveness of the ScanERU dataset, we conducted an experiment on a real-world test set. This section outlines the test set creation process and provides details about the experiment.

\begin{figure*}[tb]
 \centering
  \includegraphics[width=\textwidth]{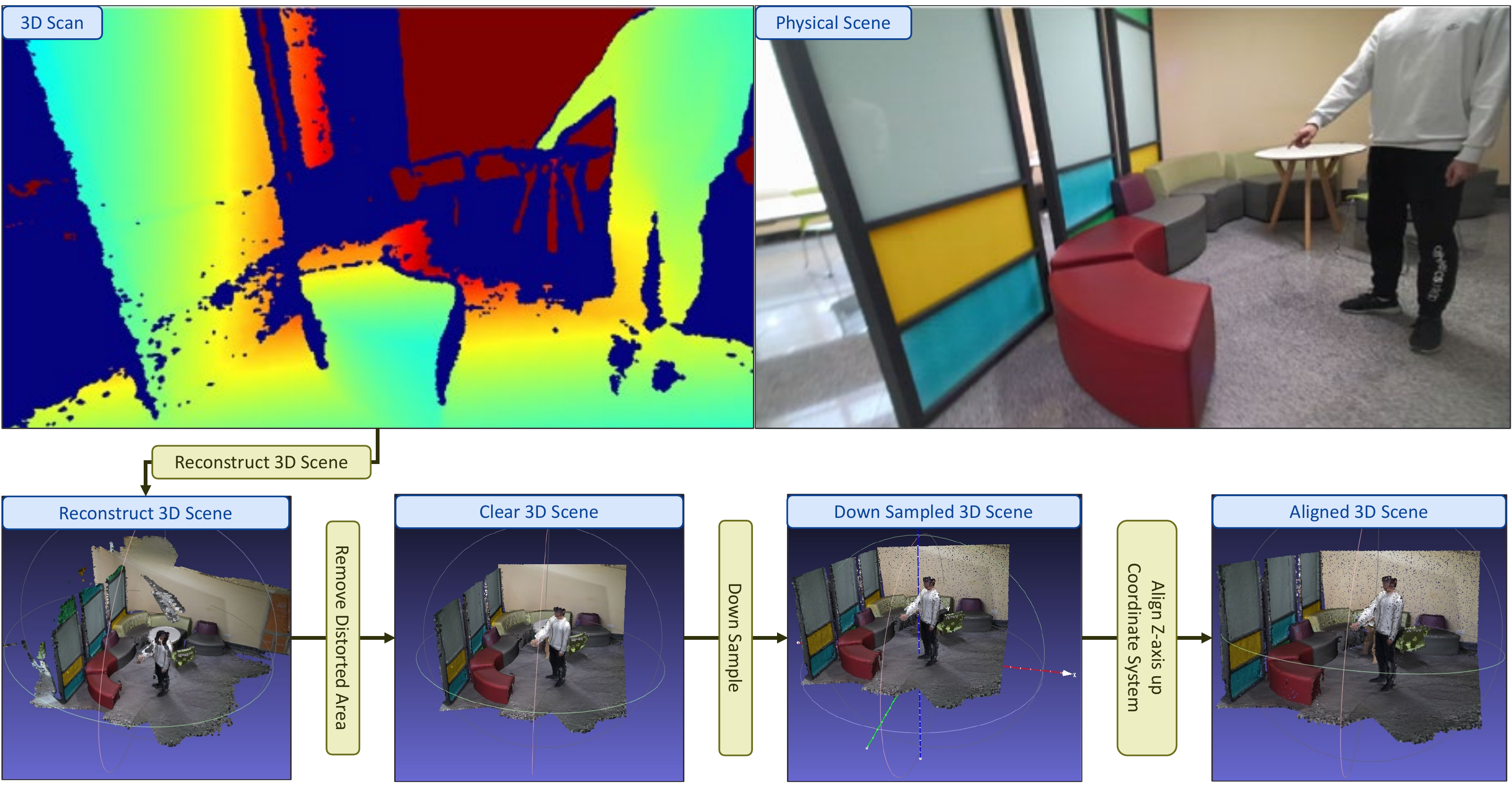}
 \caption{The data collection process. we employed the Azure Kinect DK and the official SDK, as well as the Open3D reconstruction system, to reconstruct 3D scenes. Subsequently, the 3D meshes underwent manual processing in MeshLab to eliminate distorted areas and align the point cloud to a z-axis up coordinate system.}
 \label{fig:collection}
\end{figure*}

\subsection{Data collection}

In our study, we employ the Azure Kinect DK, a Time-of-Flight (ToF) RGB-D camera with an inertial measurement unit (IMU), as our 3D sensor for reconstructing the scene. The RGB sensor operate at a resolution of 1280*540, while the depth sensor function at a resolution of 512*512. We utilize the official SDK of the Azure Kinect DK and the Open3D reconstruction system as our software. Following data acquisition, we manually process the 3D mesh data in MeshLab, by deleting distorted areas in the point cloud, especially in the edge regions, and by applying down-sampling to align the original ScanNet \cite{AngelaDai2017ScanNetR3} point cloud. Additionally, we rotate and translate the point cloud to align it to a z-axis up coordinate system. The data collection process is illustrated in ~\autoref{fig:collection}.

\subsection{Data Annotation}

\begin{figure}[tb]
 \centering
  \includegraphics[width=\columnwidth]{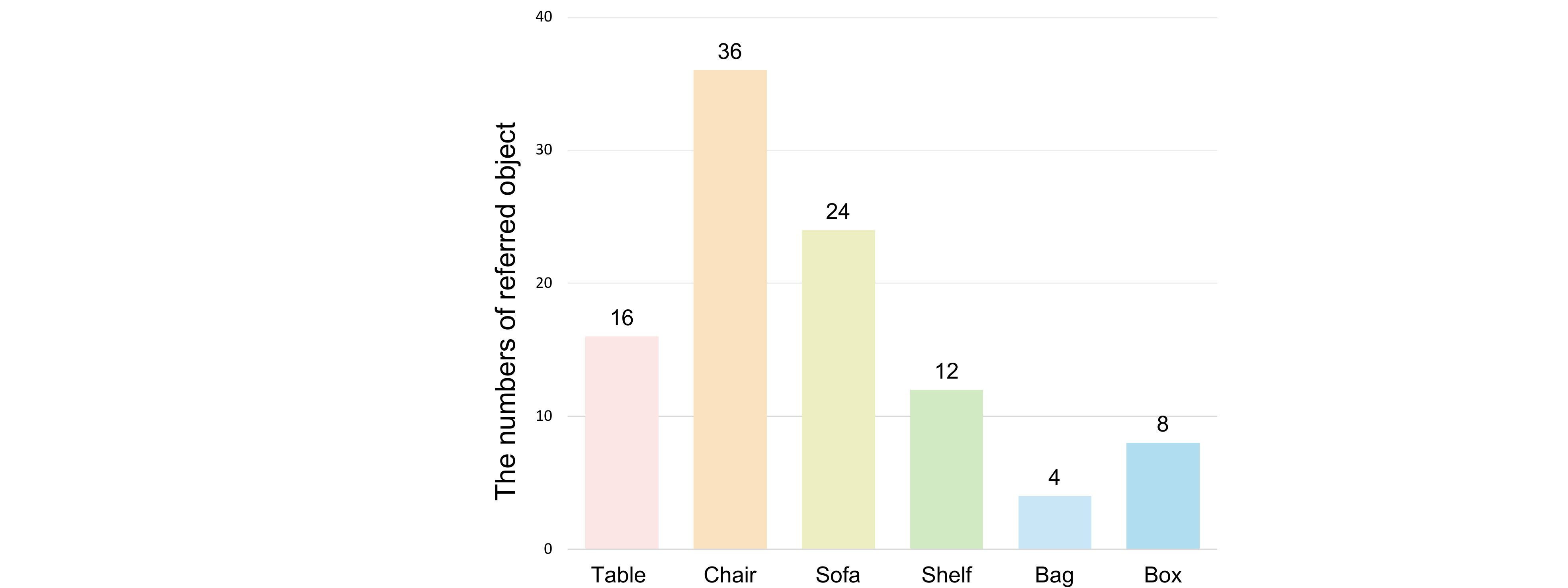}
 \caption{Dataset Statistics of test set for ScanERU.}
 \label{fig:statistics}
\end{figure}

Since these data are only used to test the validity of ScanERU, we do not perform a complete semantic segmentation annotation on the point cloud. We develope a Blender script to annotate the referred object and the human agent. For the descriptions, we apply the similar tool as ScanERU data annotation to help our workers to describe the referred object. The whole task is assigned to four workers with the request of each description (e.g. the length of description, the way to describe the referred object etc). In addition, each referred object is described by all four workers.

\subsection{Data Statistics}
In this test set, we have 100 samples. We collect 5 different scenes, consisting of 1 office, 2 lounges, and 2 classrooms, with 5 referred objects in each scene. For each referred object, we write 4 descriptions to localize them. The average length of description is 19.67 words. Following the ScanRefer \cite{chen2020scanrefer}, 100$\%$ description uses spatial language, 75$\%$ description uses color, 60$\%$ description uses shape terms, and 10$\%$ description use size information. Our human agent’s height is 176cm, following the average height of the synthetic human agents in ScanERU dataset. ~\autoref{fig:statistics} lists the distribution of the referred objects.

\subsection{Experiment}

\begin{figure*}[tb]
 \centering
  \includegraphics[width=\textwidth]{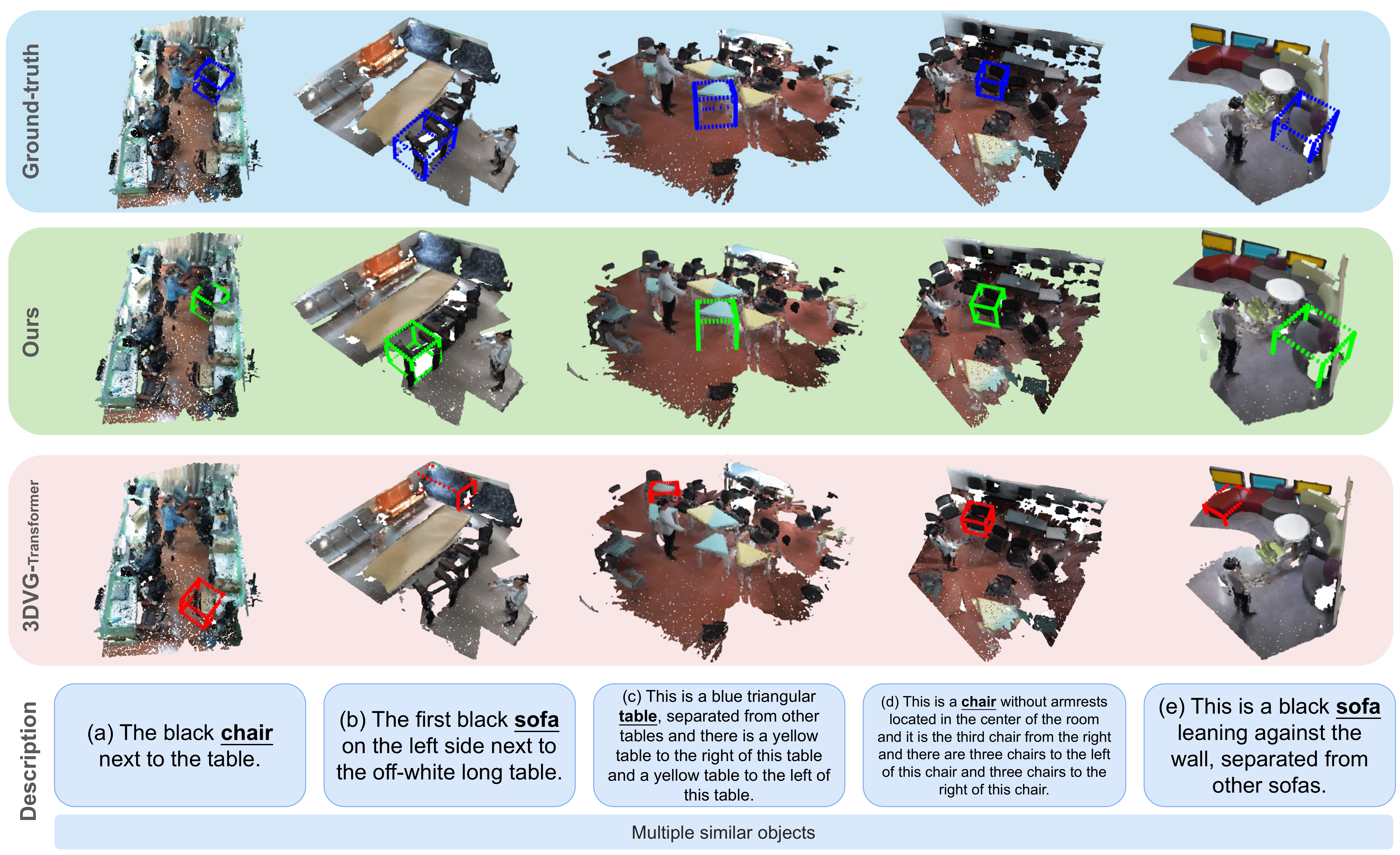}
 \caption{The localization results of 3DVG-Transformer \cite{LichenZhao20213DVGTransformerRM} and ScanERU on real-world scenes. Ground-truth bounding boxes are shown in blue, while predicted boxes are green if their IoU score with ground truth is above 50$\%$, and red otherwise. The study found their method to handle real-world environments better than 3DVG-Transformer \cite{LichenZhao20213DVGTransformerRM}, which produced inaccurate bounding boxes or failed to locate objects.}
 \label{fig:localization}
\end{figure*}

\begin{table}[b]
\caption{Comparasion of visual grounding performances on real-world test set.}

	\centering%
\begin{tabular}{cc|cc}
\hline
Methods        & Modality & Acc@0.25 & Acc@0.5 \\ \hline
ScanRefer \cite{chen2020scanrefer}  & 3D       & 34.6    & 31.0    \\
3DVG-Transformer \cite{LichenZhao20213DVGTransformerRM} & 3D       & 35.3    & 33.6   \\
Ours & 3D       & 38.0    & 34.6   \\ \hline
\end{tabular}
\label{lab:quantitive_study}
\end{table}

\textbf{Quantitive study.}
\autoref{lab:quantitive_study} shows the performance comparison of ScanRefer \cite{chen2020scanrefer}, 3DVG-Transformer \cite{LichenZhao20213DVGTransformerRM}, and ScanERU on the task of 3D object localization. We train ScanRefer \cite{chen2020scanrefer} and 3DVG-Transformer \cite{LichenZhao20213DVGTransformerRM} on the ScanRefer dataset \cite{chen2020scanrefer}, while we train our method on ScanERU dataset. Our method achieves higher accuracy than ScanRefer \cite{chen2020scanrefer} and 3DVG-Transformer \cite{LichenZhao20213DVGTransformerRM} by 3.3$\%$ and 2.6$\%$ for Acc@0.25, and 3.6$\%$ and 1.0$\%$ for Acc@0.5, respectively. The results indicate that our method is generalizable to real-world environments.

\textbf{Qualitive study.}
~\autoref{fig:localization} illustrates examples of the localization results of 3DVG-Transformer \cite{LichenZhao20213DVGTransformerRM} and ScanERU on real-world scenes. The ground-truth bounding boxes are denoted in blue, while the predicted boxes are highlighted in green if their intersection-over-union (IoU) score with the ground truth is above 50$\%$, and in red otherwise. The figure demonstrates that our method can handle complex situations in real-world environments better than 3DVG-Transformer \cite{LichenZhao20213DVGTransformerRM}, which fails to locate the correct objects or produces inaccurate bounding boxes.

\begin{figure*}[tb]
 \centering
  \includegraphics[width=\textwidth]{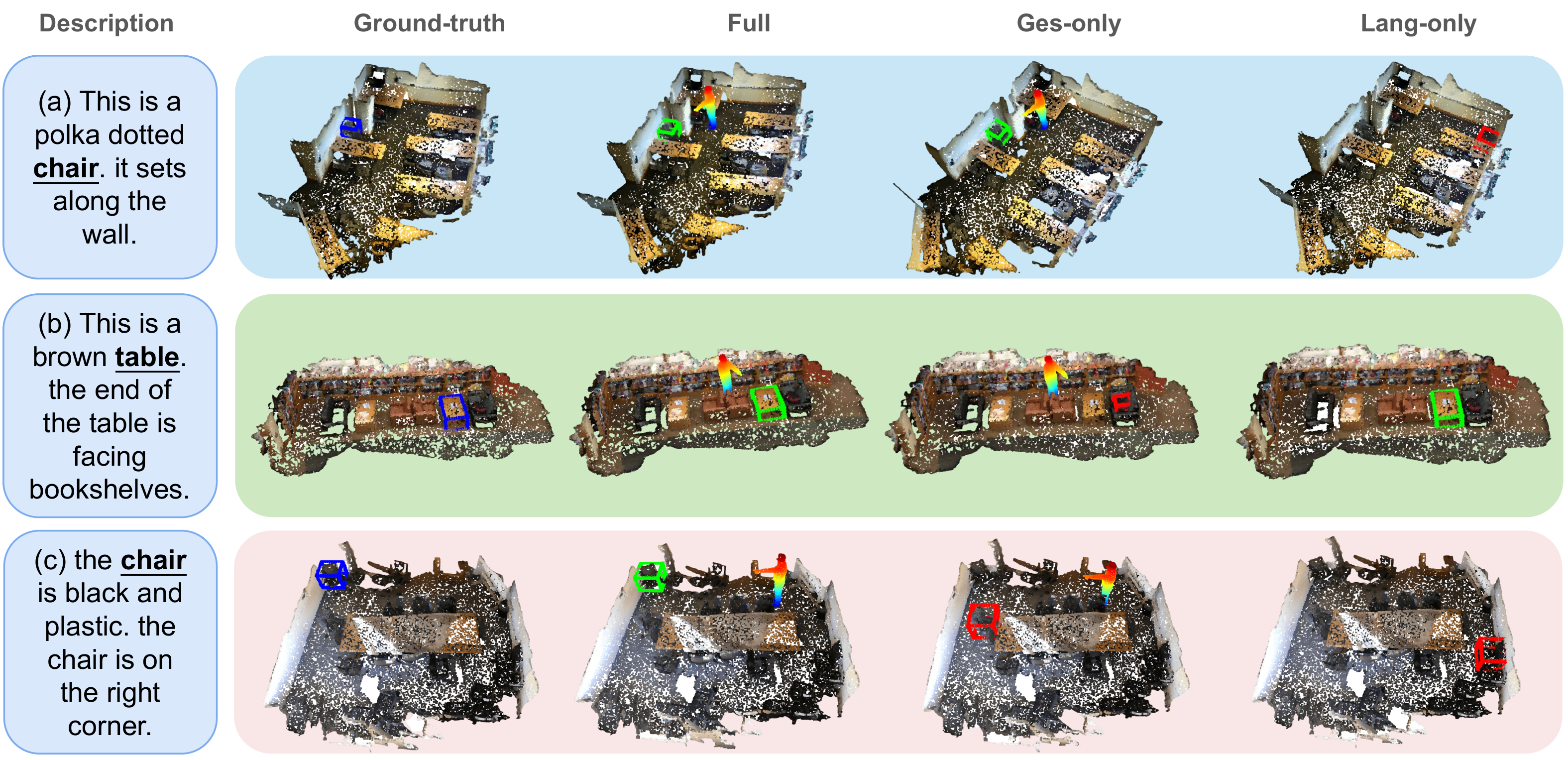}
 \caption{The localization results of the ablation study.
}
 \label{fig:ablation}
\end{figure*}

\section{Additional Qualitative Experiment for Ablation Study}
~\autoref{fig:ablation} provides a visualization of the performance of three models: Ours$_{full}$, Ours$_{lang-only}$, and Ours$_{ges-only}$. The ground-truth boxes are marked in blue, while predicted boxes with an IoU score greater than 0.5 are marked in green, otherwise they are marked in red. The results demonstrate that combining textual and gestural information is crucial for localizing objects accurately. As shown in ~\autoref{fig:ablation}(a), the description ``polka dotted chair" is challenging to understand, especially in the point cloud environment. In such cases, gesture provides easily understandable information to aid in localizing the referred object. However, as depicted in ~\autoref{fig:ablation}(b), when multiple objects, such as a sofa and table, are in the same direction pointed by the human agent, the information becomes highly ambiguous, leading to incorrect localization results. Furthermore, ~\autoref{fig:ablation}(c) demonstrates that when the textual and gestural information is ambiguous simultaneously, Ours$_{lang-only}$ and Ours$_{ges-only}$ fail to localize the referred object. In this scenario, Ours$_{full}$ can leverage both textual information, such as the word ``corner" and gestural information to disambiguate and localize the correct object and avoid the problem of potentially confusing the current perspective with the actual shooting perspective. In summary, the findings indicate that the combination of textual and gestural information is crucial for accurate object localization.

\newpage
{\small
\bibliographystyle{ieee_fullname}
\bibliography{ours}
}